\documentclass{article}
\usepackage{cite}
\usepackage{amsmath,amssymb,amsfonts}
\usepackage{algorithmic}
\usepackage{float}
\usepackage{multirow}
\usepackage{array}
\usepackage{graphicx}
\usepackage{textcomp}
\usepackage{hyperref}
\usepackage{comment}
\usepackage[left=0.75in, right=0.75in, top=0.75in, bottom=0.75in]{geometry}

\title{YOLO-LAN: Precise Polyp Detection via Optimized Loss, Augmentations and Negatives}
\author{Siddharth Gupta and Jitin Singla \\
Department of Biosciences and Bioengineering, IIT Roorkee}
\date{September 2025}

\begin{document}

\maketitle
\begin{abstract}
Colorectal cancer (CRC), a lethal disease, begins with the growth of abnormal mucosal cell proliferation called polyps in the inner wall of the colon. When left undetected, polyps can become malignant tumors. Colonoscopy is the standard procedure for detecting polyps, as it enables direct visualization and removal of suspicious lesions. Manual detection by colonoscopy can be inconsistent and is subject to oversight. Therefore, object detection based on deep learning offers a better solution for a more accurate and real-time diagnosis during colonoscopy. In this work, we propose YOLO-LAN, a YOLO-based polyp detection pipeline, trained using M2IoU loss, versatile data augmentations and negative data to replicate real clinical situations. Our pipeline outperformed existing methods for the Kvasir-seg and BKAI-IGH NeoPolyp datasets, achieving mAP$_{50}$ of 0.9619, mAP$_{50:95}$ of 0.8599 with YOLOv12 and mAP$_{50}$ of 0.9540,  mAP$_{50:95}$ of 0.8487 with YOLOv8 on the Kvasir-seg dataset. The significant increase is achieved in mAP$_{50:95}$ score, showing the precision of polyp detection. We show robustness based on polyp size and precise location detection, making it clinically relevant in AI-assisted colorectal screening.
\end{abstract}

\section{Introduction}
\label{sec:introduction}
\begin{figure}[!h]
\centerline{\includegraphics[width=0.6\columnwidth]{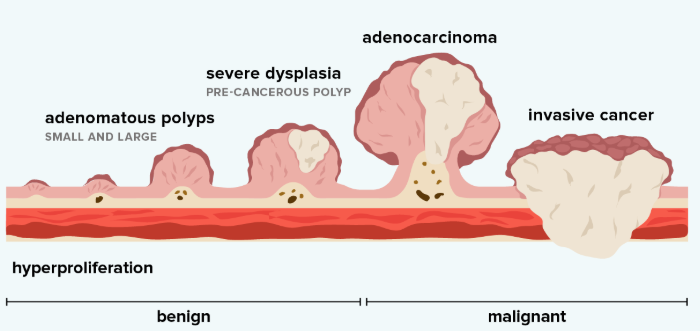}}
\caption{Schematic diagram of Benign to malignant progression of colorectal polyps adapted from \cite{a3}.}
\label{fig1}
\end{figure}

Colorectal cancer (CRC) or bowel cancer begins with abnormal growth of polyps in the rectum or colon. Although the initial growth is benign (non-cancerous), but over time, it becomes a malignant (cancerous) tumor if not detected and treated in time. The progression of polyps from benign to malignant can be seen in Fig. \ref{fig1}. As per the reports and statistics from the World Health Organization (WHO) and the Global Cancer Observatory (GLOBOCAN) 2020, colorectal cancer is ranked third on the most diagnosed cancer and second on the list of leading causes of cancer-related deaths worldwide \cite{a1}, accounting for approximately 1.9 million new cases with more than 935,000 deaths \cite{a2}. These concerning figures emphasize the importance of improved detection methods to mitigate the global risks of CRC.

Colonoscopy is the most widely used clinical method for detecting colonic polyps, although it has many limitations compared to non-invasive approaches such as virtual colonoscopy \cite{a4}. Doctors currently use colonoscopy for the detection and diagnosis of CRC. In this procedure, a lighted tube (colonoscope) with a camera attached to its tip is inserted into the colon \cite{a5}. The gastroenterologists then perform a manual examination of the colonoscopy videos to detect any polyp formation. Although colonoscopy is considered the gold standard for the detection and prevention of this disease, the manual approach is not without limitations and flaws. It takes a lot of time and is prone to observer variability with chances of high miss-rate due to fatigue or oversight. Studies have shown that polyps are often missed during colonoscopies, with miss rates ranging from 14\% to 30\%, depending on the type and size of the polyps \cite{a6} \cite{a7}\cite{a8}. While traditional Machine Learning (ML) methods have been applied to the hand-crafted features obtained from the colonoscopy and CT colonography images (example: logistic regression model on protrusion, intensity, and geometric feature \cite{a9}), it still requires domain expertise in obtaining relevant features, limiting their adaptability and generalization to diverse datasets. To address these issues, researchers have increasingly turned to end-to-end deep learning networks based on representation learning to aid in medical image analysis and diagnosis. Deep learning (DL) methods are capable of learning patterns directly from the data and make predictions \cite{a10}\cite{a11}. In the context of CRC detection, DL has been applied to recognize features of polyps in colonoscopy images, improving diagnostic consistency and speed. Convolutional neural networks (CNNs) have demonstrated exceptional performance in image recognition tasks, including medical image segmentation and object detection \cite{a12}. Object detection networks such as Faster R-CNN and SSD, the YOLO (You Only Look Once) family of models has become popular owing to thier real-time object detection capabilities, high accuracy, and efficiency \cite{a13}. 

\begin{figure}[!h]
\centerline{\includegraphics[width=0.50\columnwidth]{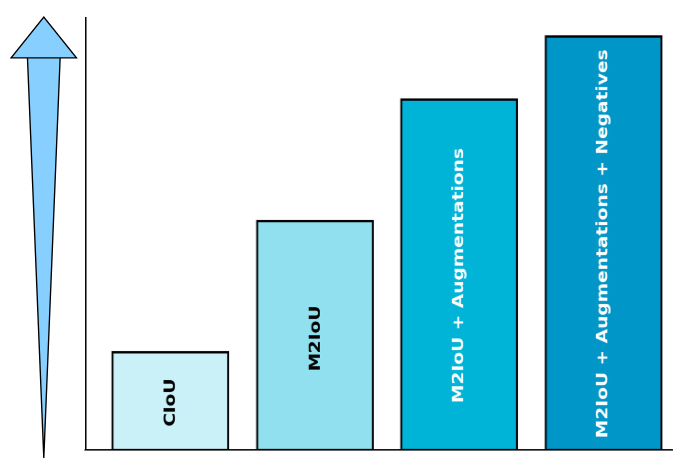}}
\caption{YOLO-LAN hypothesis illustrating increase in performance measure, especially mAP$_{50:95}$ with the inclusion of M2IoU loss, data augmentations and negative samples.}
\label{fig2}
\end{figure}

In the present work, we propose a deep learning-based pipeline called YOLO-LAN for the efficient detection of polyps in colonoscopy images. YOLO-LAN demonstrates that the inclusion of negative samples, robust augmentations, and optimized loss functions can improve the polyp detection of even small variants without architectural changes in the YOLO models.  The majority of publicly accessible datasets, such as Kvasir-seg, are naturally biased toward polyp-positive samples, which restricts the model's capacity to differentiate between polyps and polyp-look-alike regions of the colon \cite{a14}. In a realistic scenario every frame of the colonoscopy video does not contain the polyp, so the models trained on only polyp-positive samples can be biased. To overcome this issue, we introduce the negative samples from PolypGen2021 and the Kvasir datasets to the original Kvasir-seg and BKAI-IGH NeoPolyp dataset. Also, in medical imaging, the anatomical variability and acquisition artifacts are common; to make the model familiar with these conditions, data augmentations are intelligently applied. These include the spatial transformations, blur-based augmentations, and a composite mix of augmentations, which are discussed further in the paper.  We have also tested CLAHE (Contrast-Limited Adaptive Histogram Equalization) pre-processing that improves visual clarity without over-amplifying noise. Additionally, we replaced the default YOLO localization loss CIoU with M2IoU, which emphasizes the worst-case localization error by penalizing the farthest mismatched corners between predicted and ground truth boxes, resulting in significantly improved bounding box precision. Fig. \ref{fig2} illustrates the hypothesized incremental improvements: M2IoU outperforming CloU, with further gains after augmentations, and the highest improvement with inclusion of negatives.

We conducted extensive experimentation with YOLOv8 and YOLOv12 architectures, training and testing its various-sized models ranging from nano to large. For comparison with the previous studies, we randomly split the dataset into 70\% training, 20\% validation and 10\% test. The augmentations were applied only to the training samples. The augmentations on the original dataset increased its size by 10 times. We resorted to picking negative samples from the PolypGen2021 and Kvasir datasets, and these are split in the same proportion as the original dataset, and augmentations are applied only on training samples. We introduced the negative images in proportions of 10\% and 20\% of the target training data, evaluating the model’s performance in terms of its precision and specificity. We evaluate the performance of the model and report the results in terms of standard object detection metrics such as precision, recall, F1-score, and mean average precision with an IoU of 0.5 (mAP$_{50}$) and between 0.5 to 0.95 (mAP$_{50:95}$) \cite{a15}. 

YOLOv8 was specifically integrated into our proposed framework to showcase the superior performance of our pipeline over the state-of-the-art (SOTA) framework, especially mAP$_{50:95}$, as the SOTA framework used YOLOv8 in its proposed pipeline. However, as compared to the latest improved YOLO versions, YOLOv8 is a little older. Therefore, we integrated YOLOv12 into our final proposed pipeline due to its superior architectural features, lower parameter count and emphasis on a fast and advanced attention mechanism.

In summary, this work enhances the YOLO-based polyp detection on various fronts:

\begin{enumerate}
    \item   Integrating large-scale augmentation for dataset generalization within the constraints of textural and geometrical feasibility of appearance of polyps in colorectal images.
    \item  Including negative samples from colonoscopy videos to simulate real-world clinical scenarios.
    \item Incorporating M2IoU loss instead of CIoU improving the spatial alignment between the predicted and ground truth bounding boxes.
    \item Ablation study with varying negative sample ratios to assess the model’s robustness, generalization, and real-world readiness.
\end{enumerate}

The rest of the manuscript is organized as follows: Section II contains the summary of the recently published work in this field; Section III addresses the detailed methodology that contains dataset description, data augmentation and data pre-processing. It further contains the details of model architecture and the loss functions used. Section IV comprises the experiments and analysis that contain the evaluation metrics, experimental setup, ablation studies, comparison of the proposed framework with the existing literature, error analysis. Finally, Section V contains the discussion and conclusion part respectively.

\begin{table*}[ht]
\centering
\caption{Summary of the latest polyp detection studies}
\label{table1}
\renewcommand{\arraystretch}{1.2}
\setlength{\tabcolsep}{5pt}
\small
\begin{tabular}{|c|c|c|c|}
\hline
\textbf{Authors} & \textbf{Year} & \textbf{Dataset} & \textbf{Model} \\ \hline

\multirow{3}{*}{Sun et al. \cite{a16}} & \multirow{3}{*}{2025} 
 & LDPolypVideo & \multirow{3}{*}{EP-YOLO} \\ \cline{3-3}
 &  & Kvasir-SEG &  \\ \cline{3-3}
 &  & CVC-ClinicDB &  \\ \hline

\multirow{2}{*}{Sahoo et al. \cite{a17}} & \multirow{2}{*}{2025} 
 & Kvasir-SEG & YOLOv11 \\ \cline{3-4}
 &  & Kvasir-SEG (augmented) & YOLOv11 (n, s, m, l, x) \\ \hline

 Yu et al. \cite{a18} & 2025 & PolypGen & PSDM(YOLOv5 based) \\ \hline

\multirow{3}{*}{Yoo et al. \cite{a19}} & \multirow{3}{*}{2024} 
 & Kvasir-SEG & \multirow{3}{*}{YOLOv5m-TST} \\ \cline{3-3}
 &  & CVC-ClinicDB &  \\ \cline{3-3}
 &  & LDPolypVideo  &  \\ \hline

\multirow{3}{*}{Wang et al. \cite{a20}} & \multirow{3}{*}{2024} 
 & ETIS-Larib & \multirow{3}{*}{YOLO-SRPD} \\ \cline{3-3}
 &  & CVC-ClinicDB &  \\ \cline{3-3}
 &  & LDPolypVideo &  \\ \hline

\multirow{2}{*}{Khan et al. \cite{a21}} & \multirow{2}{*}{2024} 
 & Kvasir-SEG & CLAHE + YOLOv8p \\ \cline{3-4}
 &  & Hyper Kvasir-SEG & CLAHE + YOLOv8p \\ \hline

Wan et al. \cite{a22} & 2024 & Kvasir-SEG \& Gastric Polyp & CRH-YOLO \\ \hline

Ahamed et al. \cite{a23} & 2024 & Kvasir-SEG & YOLOv8 (n, s, m, l, x) \\ \hline

\multirow{5}{*}{Lalinia et al. \cite{a24}} & \multirow{5}{*}{2023} 
 & Kvasir-SEG & \multirow{5}{*}{YOLOv8 (n, s, m, l, x)} \\ \cline{3-3}
 &  & CVC-ClinicDB &  \\ \cline{3-3}
 &  & CVC-ColonDB &  \\ \cline{3-3}
 &  & ETIS &  \\ \cline{3-3}
 &  & EndoScene &  \\ \hline

\multirow{2}{*}{Nogueira et al. \cite{a25}} & \multirow{2}{*}{2023} 
 & PIBAdb cohort (train) & YOLOv3 \\ \cline{3-4}
 &  & Kvasir-SEG (test) & YOLOv3 \\ \hline


\end{tabular}
\end{table*}

\section{Related Work}
Multiple studies published over the last few years show increased performance in polyp detection on various publicly available datasets. Sun et al. in \cite{a16} proposes EP-YOLO, a lightweight model based on YOLOv10 for polyp detection. They introduced some changes in the YOLOv10 architecture by integrating a specific attention module to suppress background noise for detecting small polyps and WISE-IoU loss for improving generalization. The authors used LDPolypVideo (7,681 images), Kvasir-seg (1000 images), and CVC-ClinicDB (612 images) datasets. The proposed model shows a precision of 0.9432, a recall of 0.9117, mAP$_{50}$ of 0.9334 on the Kvasir-seg dataset. The proposed EP-YOLO outperforms models like SSD, Faster RCNN, RT-DETR, YOLOv5n, and YOLOv10n on the same dataset. Sahoo et al. \cite{a17} explored the use of an advanced YOLO architecture (YOLOv11) model for accurate and real-time colorectal polyp detection. The authors have used the Kvasir-seg dataset and the augmented Kvasir-seg dataset to train the latest YOLOv11 (nano, small, medium, large, and extra large) model. Using the augmented dataset, the precision of 0.9056, recall of 0.9514, and F1-score of 0.9186 were obtained. Overall, the recall and F1-score improved by 3.22\% and 0.93\%, respectively, but precision was reduced by 1.3\% as reported by the authors. The study demonstrates that YOLO-based models are highly effective for medical imaging and suitable for deployment in clinical settings. Yu et al.\cite{a18} recently developed a diffusion-based Progressive Spectrum Diffusion Model (PSDM) that integrates diverse clinical annotations for simultaneous detection, segmentation, and classification of polyps using the PolypGen dataset, achieving an F1-score of 0.7391, mAP$_{50}$ of 0.7624, and mAP$_{50:95}$ of 0.6083. Yoo et al. in \cite{a19} propose a lightweight transformer network YOLOv5m-TST by integrating a transformer module into the YOLOv5m backbone to enhance feature representation and capture contextual dependencies in a better way. To carry out this research, the authors used multiple datasets such as Kvasir-seg, CVC-ClinicDB, and LDPolypVideo. YOLOv5m-TST on Kvasir-seg achieved the precision of 0.9369, recall of 0.8915, F1-score 0.9134, mAP$_{50}$ of 0.9150, mAP$_{25}$ of 0.9400, and mAP$_{50}$ of 0.6855. This proposed model outperformed standard YOLOv5m and other baseline models by improving detection accuracy. Wang et al.\cite{a20} propose a novel pipeline for polyp detection named YOLO-SRPD, addressing the challenge posed by low-resolution colonoscopy images. They used Super Resolution Generative Adversarial Networks(SRGAN) for reconstructing low-resolution images into high resolution ones, enhancing textures for accurate polyp detection. These enhanced images are fed to a YOLOv5-based network integrated with attention based mixed convolutions to focus on relevant polyp features for detection. Using Kvasir-seg data, the model achieved precision of 0.867, a recall of 0.886, mAP$_{50}$ of 0.894, demonstrating competitive performance and reasonable generalization when tested cross-dataset on public datasets. Khan et al. \cite{a21} presented a real-time framework for polyp detection in endoscopy images. They applied CLAHE preprocessing and proposed a YOLOv8p polyp detector. The authors also proposed a metric named YOLO-score for assessment and performance monitoring. YOLOv8p+CLAHE achieved precision 0.962, recall 0.892, mAP$_{50}$ score of 0.955,  mAP$_{50:95}$ of 0.815 and an f1-score of 0.926 on Hyper-kvasir-seg dataset and  precision 0.952, recall 0.901, mAP$_{50}$ score of 0.945, mAP$_{50:95}$ of 0.748, and F1-score of 0.926 on Kvasir-seg dataset. Wan et al.  \cite{a22} introduced CRH-YOLO, an improved single-stage model based on YOLOv8, designed for precise and efficient detection of gastrointestinal polyps. The methodology incorporates novel modules such as CRFEM for enhanced contextual feature perception and RSPPF for improved multi-scale feature fusion, while reducing the number of detection heads and parameters to boost efficiency. The authors trained and validated their model on public datasets, including Kvasir-seg, which contains annotated images of gastric and intestinal polyps. The study demonstrated that CRH-YOLO significantly outperformed existing models, achieving key metrics such as 0.888 precision, 0.860 recall, 0.907 mAP$_{50}$ and 0.625 mAP$_{50:95}$, with notable improvements over baseline models like YOLOv8n, especially in detecting small or subtle polyps. Ahamed et al. \cite{a23} employed the YOLOv8 models (n, s, m, l, x) to train on the Kvasir-seg dataset. A subset of 100 images from the dataset was used for testing and validation. Results for each YOLOv8 variant were reported on both the test and validation sets. Among them, YOLOv8-m achieved the best overall performance, with a precision of 0.946, recall of 0.771, mAP$_{50}$ of 0.886, mAP$_{50:95}$ of 0.695, and F1-score of 0.850.   Mehrshad Lalinia and Ali Sahafi in \cite{a24} explore the application of YOLOv8, specifically its medium (YOLOv8-m) and small (YOLOv8-s) variants, for polyp detection in colonoscopy images. A custom, large-scale endoscopic dataset emphasizing low-contrast scenarios was created to train and evaluate the models. YOLOv8-m achieved impressive results with a precision of 0.956, recall of 0.917, F1-score of 0.924, mAP$_{50}$ of 0.854, and mAP$_{50:95}$ of 0.620, alongside fast inference time (10.6 ms) and 25M parameters, striking a strong balance between accuracy and speed. YOLOv8-s also performed well with slightly lower metrics but better computational efficiency. Nogueira et al. \cite{a25} retrained the YOLOv3 model that they have already published but now they injected negative samples in various proportions (2-15\%) to reduce the false positives. They used several public datasets for testing. Using the Kvasir-seg dataset, the authors report an F1-score of 0.840. The summary of the recent polyp detection work with significantly improved performance is presented in Table \ref{table1}.

\begin{figure}[ht]
\centering
\includegraphics[width=0.50\textwidth]{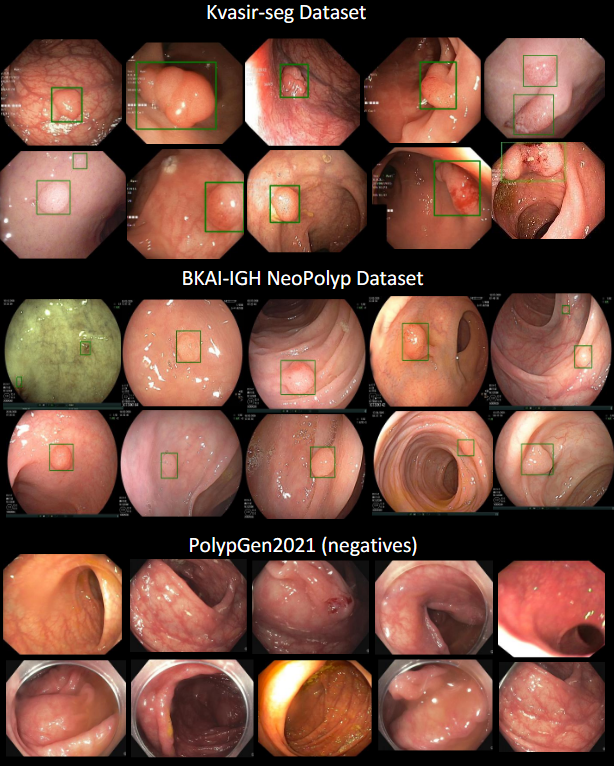}
\caption{Sample dataset images: Kvasir-seg \cite{a26} and BKAI-IGH NeoPolyp \cite{a27} datasets for polyp images, along with the ground truth bounding box and PolypGen2021 \cite{a28} dataset for negative samples.}\label{fig4}
\end{figure}

\section{Proposed Pipeline}
We have proposed a YOLO-LAN framework that consists of the following five steps: (1) Modifying the target dataset with negative samples from the Polypgen 2021 and Kvasir datasets. (2) Applying data augmentation techniques: Geometric transformations, Rotations, Blurring, Composite augmentations, (3) Splitting datasets into train, validation, and test. (4) Implementing M2IoU loss instead of default CIoU loss function for localization loss. (5) Evaluating the model (YOLOv8/12) for polyp detection task using mAP$_{50}$, mAP$_{50:95}$, precision, recall, and F1-score.

\subsection{Datasets}
The Kvasir-SEG \cite{a26} dataset is an open-access collection designed to facilitate research in gastrointestinal polyp detection and segmentation. It comprises 1,000 polyp images extracted from the Kvasir Dataset v2, each accompanied by a corresponding ground-truth segmentation mask. The images vary in resolution from 332×487 to 1920×1072 pixels and bounding box annotations for the polyps are provided in a JSON file, offering coordinate data for each image. The final masks are 1-bit depth images, where the polyp regions are represented in white (foreground), and the background is black. The Kvasir-SEG dataset is available for download at \href{https://datasets.simula.no/kvasir-seg/}{datasets.simula.no/kvasir-seg}.

The second dataset that we have used for polyp detection in our work is BKAI-IGH NeoPolyp \cite{a27} dataset. The dataset was released by the BKAI Research Center, Vietnam, comprising 1,200 endoscopic images, including 1,000 White Light Imaging (WLI) and 200 Flexible Spectral Imaging Color Enhancement (FICE) images. It is divided into a training set of 1,000 images and a test set of 200 images, with polyps annotated and classified into neoplastic (red) and non-neoplastic (green) categories. The dataset provides both segmentation and classification labels, which were verified by two expert endoscopists. In addition, a larger variant, NeoPolyp-Large, includes approximately 7,500 images captured across four imaging modalities (WLI, Blue Light Imaging (BLI), Linked Color Imaging (LCI), and FICE), offering fine-grained annotations for advanced research in computer-aided polyp detection and classification.

Additionally, to simulate realistic clinical scenarios where not all colonoscopy frames contain polyps, we incorporated 10\% and 20\% negative samples from the PolypGen2021 \cite{a28} and Kvasir \cite{a29} datasets, which consist of polyp-free colonoscopy images. While most existing methods are trained solely on polyp-containing datasets, this can lead to biased models that tend to over-detect polyps, resulting in higher false positive rates when deployed in real-world environments. By including negative samples, our model gains exposure to normal mucosal patterns, which helps reduce misclassifications and enhances robustness. This strategy not only improves generalization but also makes the model more suitable for deployment in practical clinical settings. Fig. \ref{fig4} shows the sample images from Kvasir-seg, BKAI-IGH NeoPolyp and PolypGen2021 dataset.

\subsection{Data Augmentations}

Medical image datasets are complex and often suffer from the following problems: Limited sample space, imbalanced object scales, and lighting inconsistencies. To overcome these limitations, a large amount of data is typically required to train a robust model. For this, we implemented data augmentation techniques that increased the dataset 10-fold in size, providing both the large training data as well as a regularization effect to the model.

\begin{figure*}[!t]
\centering
\includegraphics[width=1.0\textwidth]{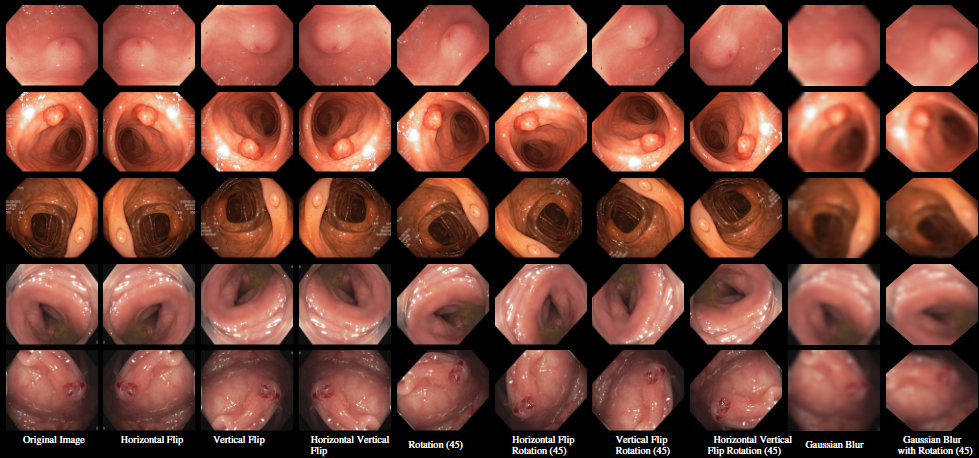}
\caption{Dataset augmentation: Original Image, Horizontal Flip, Vertical Flip, Horizontal Vertical Flip, Rotation ($45^\circ$), Horizontal Flip Rotation ($45^\circ$), Vertical Flip Rotation ($45^\circ$), Horizontal Vertical Flip Rotation ($45^\circ$), Gaussian Blur (Kernel size=15), Gaussian Blur Rotation ($45^\circ$) and CLAHE processed Image}\label{fig5}
\end{figure*}

Initially, simple spatial transformations were applied, such as horizontal flips \cite{a30}, vertical flips \cite{a30}, combined horizontal and vertical flips. These flips enable the model to recognize polyps in any orientation and develop spatial invariance because these polyps can appear on any colon wall and from multiple angles. Next, we applied rotation ($45^\circ$), both independently and with the flips. We have also included gaussian blur as a distortion augmentation approach to mitigate image loss caused by lens misalignment or rapid motion. This simulates the modest motion blur or focus loss that can be caused by rapid panning or erratic probe handling. To depict even more challenging scenarios where motion-included blur coexists with angular distortion, a composite mix of transformations such as gaussian blur with ($45^\circ$) rotation was added \cite{a31}. Some of the sample dataset images along with augmented images, are shown in Fig. \ref{fig5}.

 \subsection{Data Pre-processing}
Data-preprocessing techniques such as CLAHE \cite{a32} was applied on the training data only for the ablation study setup. CLAHE enhances image visibility by boosting local contrast in small regions, making subtle features stand out without increasing noise. Since different type of polyps such as bulbous, sessile (fold like), and flat polyps are often difficult to detect due to their low contrast against surrounding tissue \cite{a33}, applying CLAHE can be beneficial in improving their visibility and aiding detection. Fig. \ref{fig5} shows the sample images after applying CLAHE.

\subsection{Models}

Unlike traditional Computer-aided detection (CAD) approaches such as Gokturk et al. \cite{a34} which relied on statistical shape processing for specificity improvement. We used the state-of-the-art object detection YOLOv8 \cite{a35} and YOLOv12 \cite{a36} models without any architectural changes. These models developed and maintained by Ultralytics \cite{a37}, are well known for their high detection performance and inference efficiency. We implemented the YOLO-LAN pipeline with YOLOv8 for polyp detection to compare our results with YOLOv8-based state-of-the-art methods.

Compared to YOLOv8, YOLOv12 introduces a highly optimized architecture by integrating attention through Area Attention, stabilizing deep learning via R-ELAN, and simplifying computational paths via hardware-aware design choices. These innovations enable YOLOv12 to achieve state-of-the-art accuracy and speed across all model scales (n:nano, s:small, m:medium, l:large, x:extra-large), marking a significant advancement in real-time object detection and further reflecting the broader success of the convolutional neural network in medical imaging, whether trained from scratch or fine-tuned for domain-specific tasks \cite{a38}.

\subsection{Loss Function}

The loss function in YOLO models have three components: Localization loss or bounding box regression loss measures error between coordinated of ground truth box and predicted box, Objectness loss makes sure the predicted region contains some object, and Classification loss predict the correct label for the detected object. CIoU is the most popular localization loss, which takes into account three geometric factors: the Euclidean distance between the center point of the predicted bounding box and the ground truth bounding box, the difference between aspect ratios of predicted and ground truth bounding boxes and lastly, the overlap area calculated by IoU (Intersection over Union). CIoU provides better gradient flow than the original IoU loss function, resolving the vanishing gradient problem, but it does not take into consideration the distance between the corners of the actual and predicted bounding boxes. This distance becomes important for precise localization and improvement in mAP$_{50:95}$. To address this shortcoming, Shandilya et al. developed Min-max IoU (M2IoU) loss \cite{a39}. M2IoU introduces a distance-aware penalty that accelerates convergence and improves localization accuracy by treating each corner point of the bounding box independently. The brief description of the loss is as follows:

Let the ground truth bounding box be denoted by $B_{\text{gt}} = (P_1, P_2)$ and the predicted box by $B = (\hat{P}_1, \hat{P}_2)$, where $P_1$ and $P_2$ represent diagonally opposite anchor points of the box. Let $D^2(P, Q)$ denote the squared Euclidean distance between points $P$ and $Q$. Then the M2IoU loss function could be written in terms of anchor points and IoU loss as follows: 

\begin{equation}
    \mathcal{L}_{\text{M2IoU}} = 1 - \text{IoU} + \frac{\alpha D^2_{\min} + (1 - \alpha) D^2_{\max}}{C^2}
\end{equation}
where, $C$ is the diagonal length of the smallest enclosing box that covers both the predicted and ground truth boxes.  $\alpha \in [0, 1]$ is a hyperparameter controlling the penalty distribution. A smaller $\alpha$ (e.g., 0.25) imposes a greater penalty on the anchor point that is farther from the ground truth, leading to more stable and faster convergence. And lastly,
\begin{equation}
    D^2_{\max} = \max\left(D^2(P_1, \hat{P}_1), D^2(P_2, \hat{P}_2)\right)
\end{equation}
\begin{equation}
    D^2_{\min} = \min\left(D^2(P_1, \hat{P}_1), D^2(P_2, \hat{P}_2)\right)
\end{equation}
In our experiments using M2IoU consistently yielded better mAP$_{50:95}$ localization metrics compared to CIoU.
Thus, the adoption of M2IoU in our YOLO-LAN polyp detection pipeline introduces an effective enhancement tailored to the nuances of medical imaging \cite{a40}.

\begin{figure}[!htb]
\centerline{\includegraphics[width=0.5\columnwidth]{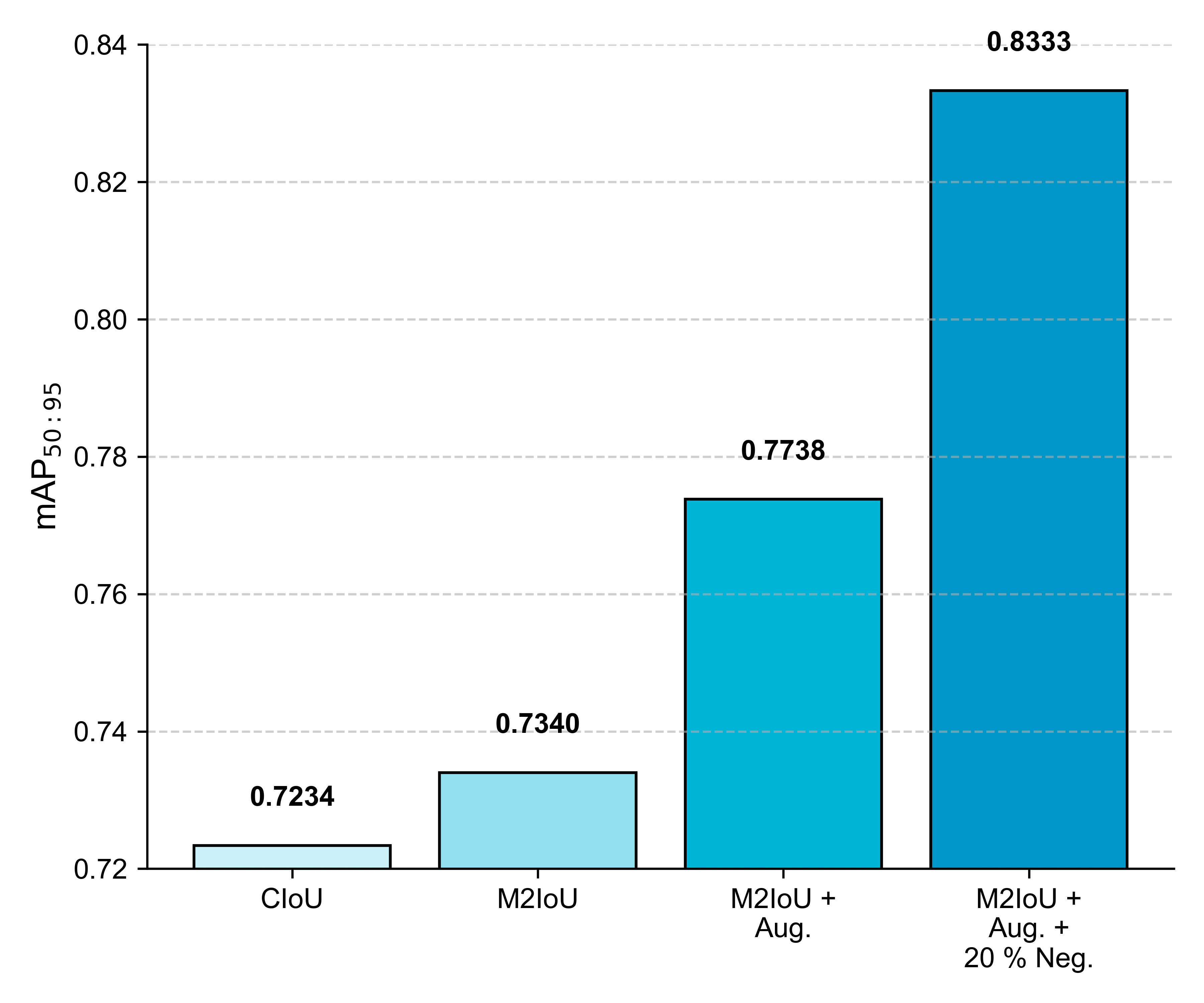}}
\caption{mAP$_{50:95}$ performance of YOLO-LAN using YOLOv8 small model on test dataset of Kvasir-Seg showing progressive improvement in polyp detection with M2IoU loss, augmentations, and negative sampling.}
\label{fig6}
\end{figure}

\begin{figure*}[!t]
\centering
\includegraphics[width=1.0\textwidth]{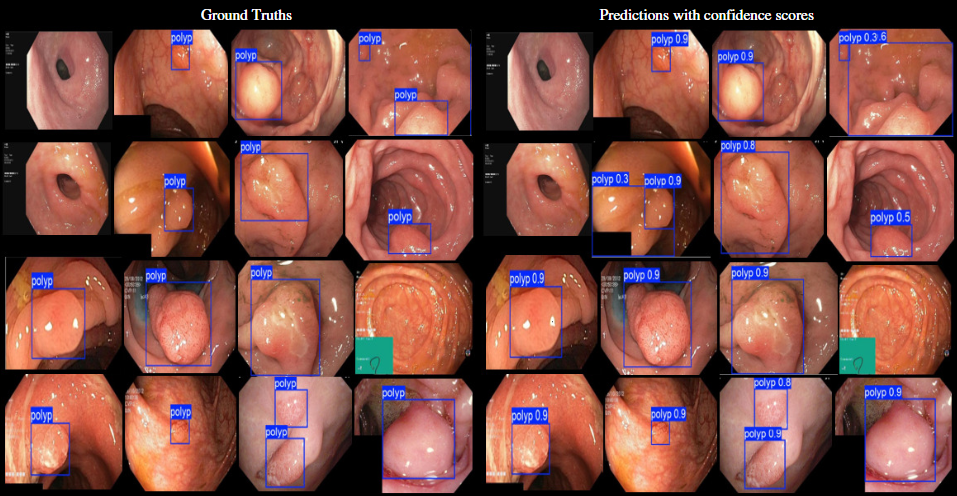}
\caption{Ground truths and Predictions with confidence scores for Kvasir-seg dataset using YOLO-LAN based on YOLOv8-small model with M2IoU loss, augmentations and 20\% negatives samples}\label{fig10}
\end{figure*}

\section{Experiments and Analysis}
In this section, a detailed description of the experiment setup, evaluation parameters, implementation details is provided. Followed by this, the results of YOLO-LAN pipeline are compared with the state-of-the-art models and results of an extensive ablation study are provided.

A series of controlled experiments are conducted to systematically evaluate the proposed pipeline under varying data conditions. For each experiment the original dataset images, along with the negative samples, are divided into 70\% training, 20\% validation, and 10\% testing sets. 

\subsection{Evaluation metrics}
To examine the effectiveness of the polyps detection task and compare with the existing methods, the following metrics are used:

\begin{itemize}
\item {Precision}: Proportion of truly predicted positive bounding boxes to the total number of predicted bounding boxes \cite{a41}. High precision means that when the model predicts an object, it's very likely to be correct, reducing false positives.

\begin{equation}
  \text{Precision} = \frac{TP}{TP + FP}
\end{equation}

\text{where:} \quad $TP$ = \text{True Positives}, \quad $FP$ = \text{False Positives}\\

\item {Recall}: Distribution of correctly positive bounding boxes to the total number of the actual positive instances \cite{a41}. The higher values of recall indicate that the model is able to detect most of the objects correctly, minimizing the number of missed detections.

\begin{equation}
  \text{Recall} = \frac{TP}{TP + FN}
\end{equation}

\text{where:} \quad $TP$ = \text{True Positives}, \quad $FN$ = \text{False Negatives}\\

\item{F1-score}: F1 Score is the harmonic mean of precision and recall. Its maximum value occurs when both the precision and recall are perfect, while its minimum value 0 occurs when either precision or recall is 0 \cite{a41}.

\begin{equation}
  \text{F1 Score} = 2 \times \frac{\text{Precision} \times \text{Recall}}{\text{Precision} + \text{Recall}}
\end{equation}

\item{mAP$_{50}$}: mAP refers to mean of Average Precision (AP) across all object classes. mAP$_{50}$ stands for mean Average Precision at IoU threshold of 50\% \cite{a42}, meaning that a predicted bounding box is considered correct if its overlap with the ground truth is at least 50\%.

\begin{equation}
  \text{mAP$_{50}$} = \frac{1}{C} \sum_{c=1}^{C} \text{AP}_c
\end{equation}

\text{where:} \quad $C$ is number of object classes.\\

\item{mAP$_{50:95}$}: mAP$_{50:95}$ stands for mean Average Precision calculated at multiple IoU thresholds ranging from 50\% to 95\% in steps of 5 \cite{a42}. The final mAP$_{50:95}$ is the average of the AP values computed across all these thresholds and all classes. mAP$_{50:95}$ is a better metric for measuring the precision of detection compared to mAP$_{50}$.

\begin{equation}
  \text{mAP$_{50:95}$} = \frac{1}{T \cdot C} \sum_{t=1}^{T} \sum_{c=1}^{C} \text{AP}_{c}^{(t)}
\end{equation}

\text{where:} \quad $C$ is number of object classes and $T$ is number of thresholds.\\
\end{itemize}

\subsection{Implementation Details}
A fixed input image resolution of 512 x 512 pixels was used to train the model on the augmentated Kvasir-seg and BKAI-IGH NeoPolyp datasets with negative samples, offering a balance between computational viability and the preservation of spatial detail. A batch size of 16 was chosen based on available GPU memory constraints. We used the Adam optimizer, which had an initial learning rate of $1 \times 10^{-4}$. Early stopping with a patience of 150 epochs was used to avoid overfitting. The model weights corresponding to the best validation performance were subsequently utilized for final testing and assessment. Table \ref{table2} shows the hardware and software configuration of the system.

\begin{table}[ht]
\centering
\caption{Hardware and Software Configurations}
\label{table2}
\setlength{\tabcolsep}{2pt}
\begin{tabular}{|p{80pt}|p{160pt}|} \hline
\textbf{Name} & \textbf{Settings} \\ \hline
Programming Language & Python 3.11.9 \\ \hline
Environment & Anaconda \\ \hline
Framework & PyTorch 2.0.1 + CUDA 12.1 \\ \hline
GPU & \par Nvidia RTX A5000 \\ & Memory: 24 GB GDDR6 with ECC \\ \hline
CPU & \par Intel(R) Xeon(R) Gold 6238R CPU 2.20GHz \\ \hline
RAM & 192 GB \\ \hline
Operating System & Windows 11 \\ \hline
\end{tabular}
\end{table}

\begin{table*}[!ht]
\centering
\caption{Comparison with literature results on Kvasir-SEG dataset}
\label{table3}
\renewcommand{\arraystretch}{1.2}
\setlength{\tabcolsep}{5pt}
\small
\begin{tabular}{|c|c|c|c|c|c|c|c|}
\hline
\textbf{Authors} & \textbf{Model} & \textbf{Precision} & \textbf{Recall} & \textbf{mAP$_{50}$} & \textbf{mAP$_{50:95}$} & \textbf{F1 score} \\ \hline
Sun et al. 2025      & EP-YOLO           & 0.9432         & 0.9117    & 0.9334         & --    & -- \\
Sahoo et al. 2025    & YOLOv11           & 0.9176          & \textbf{0.9515}& --                & --    & 0.9196 \\
Yu et al. 2025       & PSDM(YOLOv5 based)& --          &--                  & 0.7624          & 0.6083& 0.7391 \\
Yoo et al. 2024      & YOLOv5m-TST       & 0.9369          & 0.8915         & 0.9150          & 0.6855    & 0.9134 \\
Wang et al. 2024     & YOLO-SRPD         & 0.867          & 0.886          & 0.894          & --    & -- \\
Khan et al. 2024     & CLAHE + YOLOv8p   & 0.952           & 0.901           & 0.945           & 0.748 & 0.926 \\
Wan et al. 2024      & CRH-YOLO          & 0.888           & 0.860          & 0.907           & 0.625 & -- \\
Ahamed et al. 2024   & YOLOv8            & 0.946           & 0.771          & 0.886           & 0.695 & 0.850 \\
Lalinia et al. 2023  & YOLOv8m           & 0.956           & 0.917          & 0.854           & 0.620 & 0.924 \\
Nogueira et al. 2023 & YOLOv3            & --              & --             & --              & --      &0.840 \\ \hline
YOLO-LAN (Our)       & YOLOv8s (small)   & \textbf{0.9576} & 0.9231         & \textbf{0.9524} & \textbf{0.8333} & \textbf{0.9400} \\ \hline
\end{tabular}
\end{table*}

\begin{table*}[ht]
\centering
\caption{Polyp size based analysis of YOLO-LAN on Kvasir-seg test dataset}
\label{table4}
\renewcommand{\arraystretch}{1.2}
\setlength{\tabcolsep}{3pt} 
\resizebox{\textwidth}{!}{%
\begin{tabular}{|l|l|l|c|c|c|c|c|c|c|c|c|c|}
\hline
\multirow{2}{*}{\textbf{Polyp Size}} & \multirow{2}{*}{\textbf{\% of Image Area}} & \multirow{2}{*}{\textbf{\# of Polyps}} &
\multicolumn{5}{c|}{\textbf{YOLOv8s}} &
\multicolumn{5}{c|}{\textbf{YOLOv12s}} \\
\cline{4-13}
& & & \textbf{Precision} & \textbf{Recall} & \textbf{mAP$_{50}$} & \textbf{mAP$_{50:95}$} & \textbf{F1 score} 
& \textbf{Precision} & \textbf{Recall} & \textbf{mAP$_{50}$} & \textbf{mAP$_{50:95}$} & \textbf{F1 score} \\
\hline
Small  & $<$ 5    & 11 & 0.990 & 0.900 & 0.957 & 0.799 & 0.910 & 1 & 0.994 & 0.995 & 0.823 & 0.996 \\
Medium & 5--15    & 46 & 0.971 & 0.950 & 0.979 & 0.852 & 0.959 & 1 & 0.923 & 0.980 & 0.865 & 0.959 \\
Large  & $>$ 15   & 53 & 0.942 & 0.763 & 0.827 & 0.728 & 0.842 & 0.893 & 0.797 & 0.839 & 0.747 & 0.842 \\
\hline
\end{tabular}%
}
\end{table*}

\subsection{Effect of Loss, Augmentations and Negatives}

In the first experiment, we trained the YOLOv8-s (small, 11.2 million parameters) model on the original Kvasir-seg dataset. The model was trained under various incremental conditions such as with conventional CIoU loss, with M2IoU loss, and subsequently adding data augmentations and 10\% and 20\% of negative samples.

Our results show that mAP$_{50:95}$ undergoes a significant increase with each modification. In Fig. \ref{fig6}, we can see that for test dataset of Kvasir-seg, the mAP$_{50:95}$ increases from 0.7234 with CIoU to 0.7340 with M2IoU loss. It further increases to 0.7738 with data augmentation and to 0.8333 with 20\% negatives. Overall, our pipeline showed an increase of 15.19\% in mAP$_{50:95}$ compared to baseline YOLOv8-small model with CIoU without augmentations and negatives. This aligns with our hypothesis in Fig. \ref{fig2} that YOLO-LAN increases precision of predicted polyp localization. Similarly, mAP$_{50}$, precision and recall showed an increase of 8.73\%, 5.81\% and 10.51\%, respectively, from CIoU to M2IoU + Augmentations + 20\% Negatives. Fig. \ref{fig10} shows the ground truths and the predictions with confidence scores for Kvasir-seg dataset using YOLO-LAN based on YOLOv8-small with M2IoU loss, augmentations and 20\% negatives samples.

\subsection{Comparison with the state-of-the-art models}

Numerous studies have been published in the last 5 years on polyp detection testing methods on a variety of datasets. Although most of the methods showed improved performance of mAP$_{50}$, they lack significant improvement in mAP$_{50:95}$.
We compare YOLO-LAN with available state-of-the-art results from the literature on Kavasir-seg dataset. YOLO-LAN with YOLOv8-s model, M2IoU loss, augmentations, and 20\% negative samples outperformed all existing methods, reaching 0.9576 precision, 0.9231 recall, 0.9400 F1-score, 0.9524 mAP$_{50}$ and 0.8333 mAP$_{50:95}$ (Table \ref{table3}). Most models with more than 90\% mAP$_{50}$ falls short of the performance on mAP$_{50:95}$. But YOLO-LAN bridges that gap. Methods reported in Table \ref{table3} already outperform many other previously reported results, pointing the significant improvement in localization precision of YOLO-LAN in polyp detection. This shows that the YOLO-LAN workflow with vanilla architecture can outperform other methods with special architectural changes designed for polyp detection.

For BKAI-IGH NeoPolyp dataset, we provide a new benchmark for polyp detection as no studies on detection were available in the literature. With YOLOv8-s we achieved mAP$_{50:95}$ of 0.7984 and mAP$_{50}$ of 0.8982 as shown in Table \ref{table5}.

\subsection{Evaluating Model Sensitivity to Polyp Size}
We also tested the performance of YOLO-LAN on different sizes of the polyps within the Kvasir-seg test dataset. Overall improvement in localization may not necessarily mean that the model performance across polyp size is precise. To test this, we divided the test dataset into three classes based on the bounding box area relative to the image area. Small size polyps with a bounding box area smaller than 5\%, medium size polyps with an area between 5-15\%, and large size polyps with more than 15\% bounding box area. The findings shown in Table \ref{table4} suggest that YOLO-LAN has relatively high precision and recall for small and medium sized polyps with F1-scores at least 0.95 and mAP$_{50:95}$ more than 0.80. The performance for detecting large polyps was relatively poor compared to overall test performance in Table \ref{table4}, especially recall and mAP values. This suggests the model is very good at localizing smaller and medium polyps, which are relatively more challenging to detect manually in real-world scenarios.

\begin{table*}[ht]
\centering
\caption{Performance of YOLOv8 variants on Kvasir-seg and BKAI-IGH NeoPolyp datasets}
\label{table5}
\renewcommand{\arraystretch}{1.2}
\setlength{\tabcolsep}{4pt} 
\resizebox{\textwidth}{!}{%
\begin{tabular}{|l|l|c|c|c|c|c|@{\hskip 0pt}|c|c|c|c|c|}
\hline
\multirow{2}{*}{\textbf{Experiment}} & \multirow{2}{*}{\textbf{Model}} &
\multicolumn{5}{c|}{\textbf{Kvasir-seg}} &
\multicolumn{5}{c|}{\textbf{BKAI-IGH NeoPolyp }} \\
\cline{3-12}
 & & \textbf{Precision} & \textbf{Recall} & \textbf{mAP$_{50}$} & \textbf{mAP$_{50:95}$} & \textbf{F1 score} 
   & \textbf{Precision} & \textbf{Recall} & \textbf{mAP$_{50}$} & \textbf{mAP$_{50:95}$} & \textbf{F1 score} \\
\hline
\multirow{4}{*}{Baseline datasets (CIoU)} 
& YOLOv8l & 0.8938 & 0.8333 & 0.8448 & 0.6929 & 0.8625 & 0.8596 & 0.9000 & 0.8131 & 0.6937 & 0.8793 \\
& YOLOv8m & 0.9017 & 0.8421 & 0.8601 & 0.7133 & 0.8709 & 0.8972 & 0.9000 & 0.8112 & 0.7105 & 0.8986 \\
& YOLOv8s & 0.9050 & 0.8353 & 0.8759 & 0.7234 & 0.8688 & 0.9124 & 0.8929 & 0.8431 & 0.7279 & 0.9025 \\
& YOLOv8n & 0.9418 & 0.8246 & 0.8738 & 0.7301 & 0.8793 & 0.7893 & 0.9634 & 0.8231 & 0.6961 & 0.8677 \\
\hline
\multirow{4}{*}{Baseline datasets (M2IoU)} 
& YOLOv8l & 0.9270 & 0.7982 & 0.8667 & 0.7046 & 0.8578 & 0.8596 & 0.9000 & 0.8135 & 0.6942 & 0.8793 \\
& YOLOv8m & 0.8810 & 0.7796 & 0.8320 & 0.6694 & 0.8272 & 0.8972 & 0.9000 & 0.8095 & 0.7092 & 0.8986 \\
& YOLOv8s & 0.8938 & 0.8421 & 0.8678 & 0.7340 & 0.8672 & 0.9126 & 0.8947 & 0.8442 & 0.7272 & 0.9035 \\
& YOLOv8n & 0.9201 & 0.7895 & 0.8736 & 0.7518 & 0.8498 & 0.8130 & 0.9571 & 0.8198 & 0.6938 & 0.8792 \\
\hline
\multirow{4}{*}{M2IoU + Augmented data} 
& YOLOv8l & 0.9062 & 0.8472 & 0.8982 & 0.7954 & 0.8757 & 0.8042 & 0.9429 & 0.8598 & 0.7515 & 0.8680 \\
& YOLOv8m & 0.9592 & 0.8244 & 0.8992 & 0.7961 & 0.8867 & 0.8342 & 0.9714 & 0.8833 & 0.7396 & 0.8976 \\
& YOLOv8s & 0.9143 & 0.8420 & 0.8869 & 0.7738 & 0.8767 & 0.8431 & 0.9571 & 0.8919 & 0.7945 & 0.8965 \\
& YOLOv8n & 0.9201 & 0.7895 & 0.8736 & 0.7518 & 0.8498 & 0.8680 & 0.9714 & 0.8950 & 0.7847 & 0.9168 \\
\hline
\multirow{4}{*}{\shortstack{M2IoU + Augmented data + \\ 10\% Negative data}} 
& YOLOv8l & 0.8862 & 0.8509 & 0.8898 & 0.7983 & 0.8626 & 0.8130 & 0.9317 & 0.8746 & 0.7648 & 0.8683 \\
& YOLOv8m & \textbf{0.9597} & 0.8158 & 0.8778 & 0.7622 & 0.8819 & 0.8035 & \textbf{0.9857} & 0.8799 & 0.7783 & 0.8854 \\
& YOLOv8s & 0.8817 & 0.8498 & 0.8734 & 0.7534 & 0.8654 & \textbf{0.9177} & 0.9429 & 0.8601 & 0.7555 & \textbf{0.9301} \\
& YOLOv8n & 0.9133 & 0.8321 & 0.8677 & 0.7598 & 0.8708 & 0.8419 & 0.9571 & \textbf{0.9017} & 0.7766 & 0.8958 \\
\hline
\multirow{4}{*}{\shortstack{M2IoU + Augmented data + \\ 20\% Negative data}} 
& YOLOv8l & 0.9442 & 0.8558 & 0.9402 & 0.8392 & 0.8978 & 0.8130 & 0.9317 & 0.8746 & 0.7648 & 0.8683 \\
& YOLOv8m & 0.9027 & 0.9038 & 0.9366 & 0.8369 & 0.9033 & 0.8787 & 0.9312 & 0.8967 & 0.7922 & 0.9042 \\
& YOLOv8s & 0.9576 & 0.9231 & 0.9524 & 0.8333 & \textbf{0.9400} & 0.7899 & 0.9667 & 0.8982 & \textbf{0.7984} & 0.8694 \\
& YOLOv8n & 0.9010 & 0.9038 & 0.9412 & 0.8174 & 0.9024 & 0.8490 & 0.9286 & 0.8660 & 0.7632 & 0.8870 \\
\hline
\multirow{4}{*}{\shortstack{M2IoU + Augmented data + \\ 20\% Negative data + \\ CLAHE}} 
& YOLOv8l & 0.9573 & 0.9231 & \textbf{0.9540} & \textbf{0.8487} &0.9399 & 0.8572 & 0.9437 & 0.8404 & 0.7359 & 0.8983 \\
& YOLOv8m & 0.9243 & \textbf{0.9395} & 0.9429 & 0.8232 & 0.9319 & 0.8760 & 0.9296 & 0.8376 & 0.7360 & 0.9020 \\
& YOLOv8s & 0.9104 & 0.9038 & 0.9303 & 0.8055 & 0.9071 & 0.7893 & 0.9497 & 0.8192 & 0.7209 & 0.8621 \\
& YOLOv8n & 0.9164 & 0.8654 & 0.9389 & 0.8070 & 0.8902 & 0.8095 & 0.9296 & 0.8190 & 0.7217 & 0.8654 \\
\hline
\end{tabular}%
}
\end{table*}

\begin{table*}[ht]
\centering
\caption{Performance of YOLOv12 variants on Kvasir-seg and BKAI-IGH NeoPolyp datasets}
\label{table6}
\renewcommand{\arraystretch}{1.2}
\setlength{\tabcolsep}{3pt} 
\resizebox{\textwidth}{!}{%
\begin{tabular}{|l|l|c|c|c|c|c|@{\hskip 0pt}|c|c|c|c|c|}
\hline
\multirow{2}{*}{\textbf{Experiment}} & \multirow{2}{*}{\textbf{Model}} &
\multicolumn{5}{c|}{\textbf{Kvasir-seg}} &
\multicolumn{5}{c|}{\textbf{BKAI-IGH NeoPolyp }} \\
\cline{3-12}
 & & \textbf{Precision} & \textbf{Recall} & \textbf{mAP$_{50}$} & \textbf{mAP$_{50:95}$} & \textbf{F1 score} 
   & \textbf{Precision} & \textbf{Recall} & \textbf{mAP$_{50}$} & \textbf{mAP$_{50:95}$} & \textbf{F1 score} \\
\hline
\multirow{4}{*}{Baseline datasets (CIoU)} 
& YOLOv12l & 0.9380 & 0.8158 & 0.8587 & 0.7057 & 0.8726 & 0.7369 & 0.9429 & 0.7863 & 0.6796 & 0.8273 \\
& YOLOv12m & 0.8662 & 0.7948 & 0.8275 & 0.6938 & 0.8290 & 0.8164 & 0.9143 & 0.8302 & 0.6870 & 0.8626 \\
& YOLOv12s & 0.8929 & 0.8333 & 0.8580 & 0.7284 & 0.8621 & 0.8362 & 0.9286 & 0.8385 & 0.7384 & 0.8799 \\
& YOLOv12n & 0.9121 & 0.8196 & 0.8673 & 0.7185 & 0.8634 & \textbf{0.9027} & 0.8429 & 0.7986 & 0.6771 & 0.8717 \\
\hline
\multirow{4}{*}{Baseline datasets (M2IoU)} 
& YOLOv12l & 0.9285 & 0.8735 & 0.9384 & 0.7831 & 0.9001 & 0.7346 & 0.9429 & 0.7860 & 0.6785 & 0.8258 \\
& YOLOv12m & 0.9226 & 0.8446 & 0.9313 & 0.7903 & 0.9032 & 0.8175 & 0.9143 & 0.8302 & 0.6867 & 0.8632 \\
& YOLOv12s & 0.9107 & 0.8846 & 0.9417 & 0.7827 & 0.8989 & 0.8362 & 0.9286 & 0.8394 & 0.7390 & 0.8799 \\
& YOLOv12n & 0.9037 & 0.8942 & 0.9223 & 0.7794 & 0.8990 & 0.9026 & 0.8429 & 0.7996 & 0.6794 & 0.8717 \\
\hline
\multirow{4}{*}{M2IoU + Augmented data} 
& YOLOv12l & \textbf{0.9688} & 0.8509 & 0.9143 & 0.8187 & 0.9060 & 0.8439 & 0.9268 & 0.8511 & 0.7260 & 0.8834 \\
& YOLOv12m & 0.9504 & 0.8400 & 0.8848 & 0.7594 & 0.8918 & 0.8597 & 0.9287 & 0.8634 & 0.7421 & 0.8928 \\
& YOLOv12s & 0.9582 & 0.8158 & 0.9062 & 0.8034 & 0.8813 & 0.8486 & 0.9286 & 0.8523 & 0.7363 & 0.8868 \\
& YOLOv12n & 0.9460 & 0.8333 & 0.8640 & 0.7300 & 0.8861 & 0.8742 & 0.9286 & 0.8336 & 0.7280 &0.9006 \\
\hline
\multirow{4}{*}{\shortstack{M2IoU + Augmented data + \\ 10\% Negative data}} 
& YOLOv12l & 0.9417& 0.8505 & 0.9004 & 0.7949 & 0.8938 & 0.8049 & 0.9571 & 0.8762 & 0.7499 & 0.8744 \\
& YOLOv12m & 0.8972& 0.8509 & 0.8941 & 0.7912 & 0.8734 & 0.8862 & 0.8897 & 0.8613 & 0.7467 & 0.8880 \\
& YOLOv12s & 0.9347& 0.8509 & 0.9149 & 0.8133 & 0.8908 & 0.8010 & 0.9286 & 0.8590 & 0.7274 & 0.8601 \\
& YOLOv12n & 0.9141& 0.8772 & 0.8963 & 0.7973 & 0.8953 & 0.8330 & 0.9571 & 0.8659 & 0.7543 & 0.8908 \\
\hline
\multirow{4}{*}{\shortstack{M2IoU + Augmented data + \\ 20\% Negative data}} 
& YOLOv12l & 0.8973 & 0.9231 & 0.9575 & \textbf{0.8587} & 0.9100 & 0.8794 & 0.9377 & \textbf{0.8846} & \textbf{0.7567} & \textbf{0.9076} \\
& YOLOv12m & 0.9496 & 0.9231 & 0.9611 & 0.8578 & 0.9362 & 0.8229 & \textbf{0.9571} & 0.8586 & 0.7390 & 0.8849 \\
& YOLOv12s & 0.9166 & 0.9507 & 0.9581 & 0.8413 & 0.9333 & 0.8331 & 0.9570 & 0.8538 & 0.7447 & 0.8908 \\
& YOLOv12n & 0.9483 & 0.9135 & \textbf{0.9619} & 0.8398 & 0.9306 & 0.8582 & 0.9286 & 0.8271 & 0.7126 & 0.8920 \\
\hline
\multirow{4}{*}{\shortstack{M2IoU + Augmented data + \\ 20\% Negative data + \\ CLAHE}} 
& YOLOv12l & 0.8974 & 0.9231 & 0.9532 & 0.8428 & 0.9101 & 0.8572 & 0.9437 & 0.8404 & 0.7359 & 0.8983 \\
& YOLOv12m & 0.9033 & 0.9135 & 0.9434 & 0.8231 & 0.9083 & 0.7786 & 0.8917 & 0.8285 & 0.7173 & 0.8313 \\
& YOLOv12s & 0.9409 & \textbf{0.9519} & 0.9543 & 0.8437 & \textbf{0.9464} & 0.8252 & 0.8592 & 0.8116 & 0.7100 & 0.8418 \\
& YOLOv12n & 0.9213 & 0.8558 & 0.9429 & 0.8319 & 0.8873 & 0.8431 & 0.9014 & 0.8360 & 0.7377 & 0.8713 \\
\hline
\end{tabular}
}
\end{table*}

\subsection{Ablation Study}

Here we report the results of ablation studies performed on YOLO-LAN over YOLOv8 and YOLOv12 model sizes, percentage of negative samples and preprocessing techniques like CLAHE. We trained nano, small, medium and large YOLOv8/12 models on both datasets with CIoU, M2IoU, + data augmentation, + 10\% or 20\% negative samples. YOLOv12 was used to test the performance of YOLO-LAN pipeline with latest YOLO models.

Starting with a baseline setup where we trained nano, small, medium, and large variants of YOLOv8 on both datasets independently. Baseline setup excludes any data pre-processing, data augmentation, or negative samples. With the original train dataset that includes 700 polyp images, the baseline YOLOv8 small model with CIoU achieves best value of mAP$_{50}$ = 0.8759 and YOLOv8 nano model achieves the best value of mAP$_{50:95}$ = 0.7301. When all the augmentations are applied on the train dataset along with M2IoU loss, the substantial improvement is observed in mAP$_{50:95}$ = 0.7961 using YOLOv8 medium variant. Further with the inclusion of 10\% negatives (70 samples), we observe a slight drop in the mAP$_{50}$ = 0.8898, but improved robustness as indicated by the stable mAP$_{50:95}$ = 0.7983. Increasing the number of negative samples to 20\% achieved mAP$_{50}$ and mAP$_{50:95}$ scores of 0.9524 from YOLOv8 small and 0.8392 from YOLOv8 large variants, respectively. The results stated in Table \ref{table5} indicate the model's discriminative ability, even if minor falls in detection scores occur. Finally, we tested CLAHE pre-processing as previous studies report model performance improvement with CLAHE. However, CLAHE preprocessing did not improve the performance of small model, but did show some improvement in mAP$_{50:95}$ and mAP$_{50}$ with large model. The marginal mAP$_{50}$ improvement of 0.0016 achieved with CLAHE in the large model may not justify the substantial increase in parameter size from 11.2 million to 43.7 million compared to the small model without CLAHE \cite{a35}.

We observed that small models with addition of negative samples are able to show similar performance as large models without negatives, for example, YOLOv8 large has mAP$_{50}$ of 0.8982 with M2IoU and data augmentations along, whereas, YOLOv8 small has mAP$_{50}$ of 0.9524 with 20\% negatives.

Recognizing that YOLOv8 is now a previous-generation model, we also evaluated YOLO-LAN with latest YOLOv12 model. Same set of experiments were conducted for ablation study as with YOLOv8. As shown in Table \ref{table6}, YOLOv12 nano with M2IoU, data augmentations and 20\% negative samples on the Kvasir-seg dataset shows the overall best performance with highest mAP$_{50}$ of 0.9619 with YOLOv12 nano and high mAP$_{50:95}$ of 0.8587 by YOLOv12 large. These scores set another benchmark for the poly detection.

The results on BKAI-IGH NeoPolyp dataset also demonstrate a strong upward trend in detection performance by applying YOLO-LAN pipeline. Starting from lower baselines, the incorporation of optimized data augmentation, a well-balanced mix of negative samples, and improved training approaches led to gains across all evaluation metrics as shown in Table \ref{table5} and \ref{table6} for YOLOv8 and YOLOv12 respectively. With YOLOv12, highest recall, mAP$_{50}$, mAP$_{50:95}$ and F1-score is achieved with M2IoU, data augmentations and 20\% negatives. These findings confirm the effectiveness of the proposed YOLO-LAN pipeline in significantly boosting performance for polyp detection.

\subsection{Evaluation and Testing of Candidate Models}
\begin{figure*}[t!]
    \centering
    \includegraphics[width=1.0\textwidth]{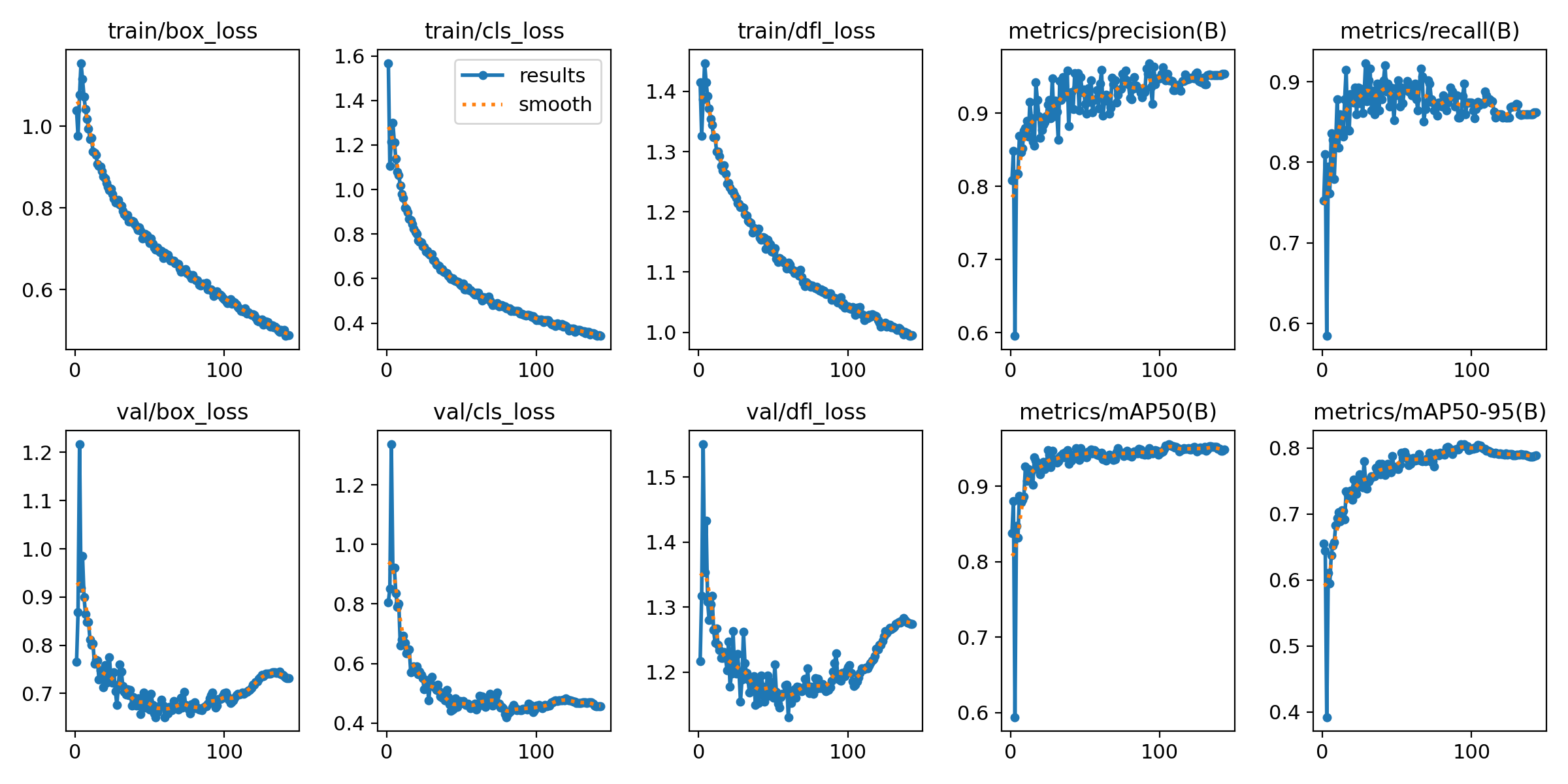}
    \caption{Training and validation behavior of YOLO-LAN pipeline with YOLOv8s model, M2IoU localization loss, data augmentations and 20\% negative samples. The behavious is shown for 200 epochs on Kvasir-seg dataset. Metrics include box loss, classification loss, DFL loss, precision, recall, mAP$_{50}$, and mAP$_{50:95}$.}
    \label{fig7}
\end{figure*}

\begin{figure*}[t!]
    \centering
    \includegraphics[width=1.0\textwidth]{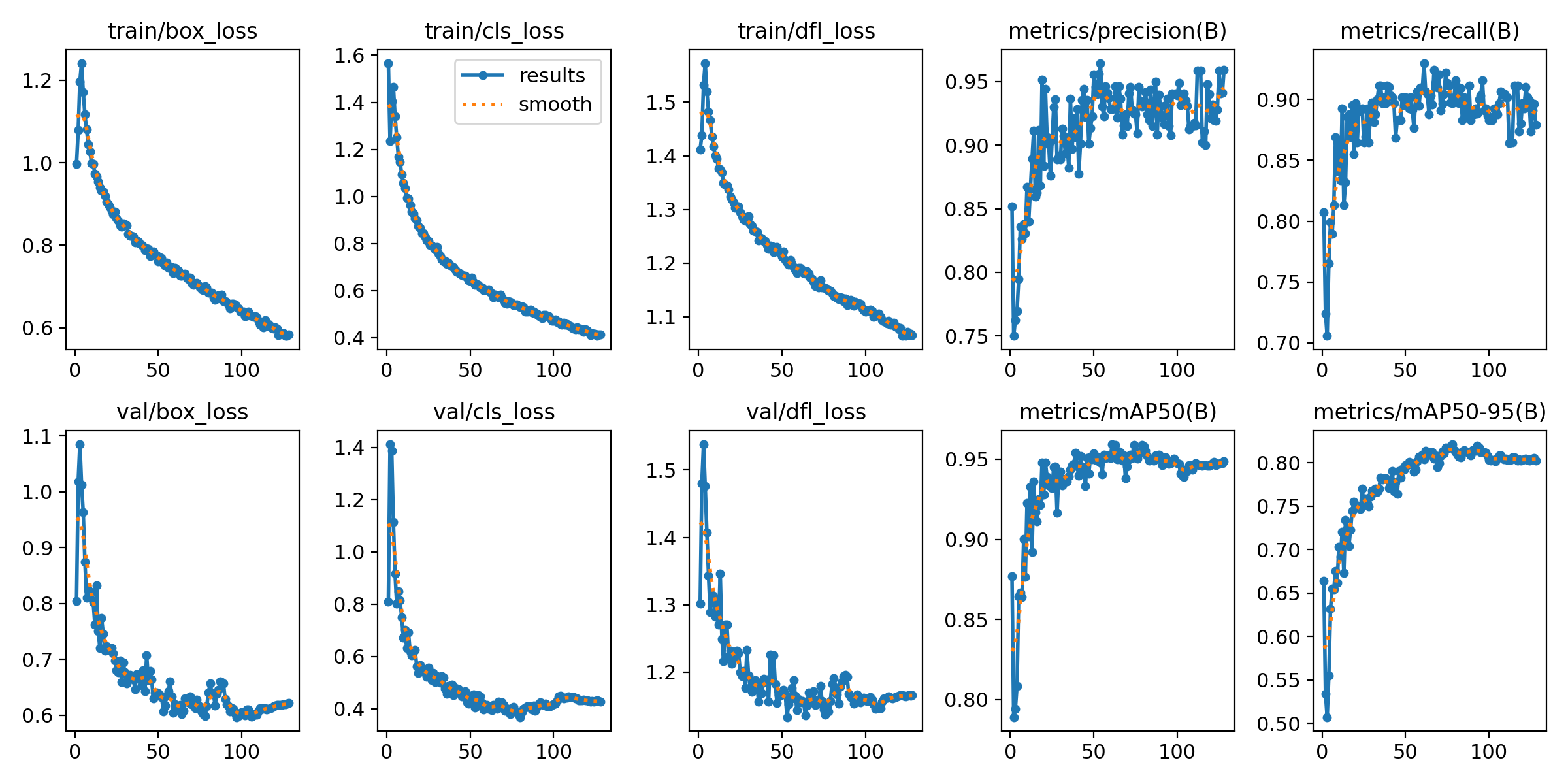}
    \caption{Training and validation behavior of YOLO-LAN pipeline with YOLOv12s model, M2IoU localization loss, data augmentations and 20\% negative samples. The behavious is shown for 200 epochs on Kvasir-seg dataset. Metrics include box loss, classification loss, DFL loss, precision, recall, mAP$_{50}$, and mAP$_{50:95}$.}
    \label{fig8}
\end{figure*}

Fig. \ref{fig7} and \ref{fig8} represents the evolution of training and validation losses along with evaluation metrics across 200 epochs for the YOLOv8 small and YOLOv12 small model trained on the Kvasir-seg dataset. In both the figures the training loss (box loss, cls loss, dfl loss) decreases smoothly, indicating stable convergence. However, when we see the validation loss, we observe the same trends only minor oscillations can be observed in DFL loss. YOLOv12 small attain the higher values for precision, recall, and mAP$_{50:95}$ in comparison to the YOLOv8 small model. The overall figures shows that the YOLOv12 provides good generalization and improved detection performance, especially under stricter IoU thresholds, demonstrating its architectural advantage over YOLOv8.

\subsection{Error Analysis}
We conducted an error analysis to check the limitations of our models. We begin by categorizing the test set polyp images into the small, medium, and large categories as discussed in the previous subsection, and for each image, we overlay bounding boxes with predictions in red and ground truths in green. This analysis enabled a clear visualization of detection discrepancies, where failure cases such as over and under extended bounding boxes, missing detections, and inaccurate localization were identified across different polyp size categories. The detailed observations for each class were described below:

\begin{figure}[H]
\centering
\includegraphics[width=0.50\textwidth]{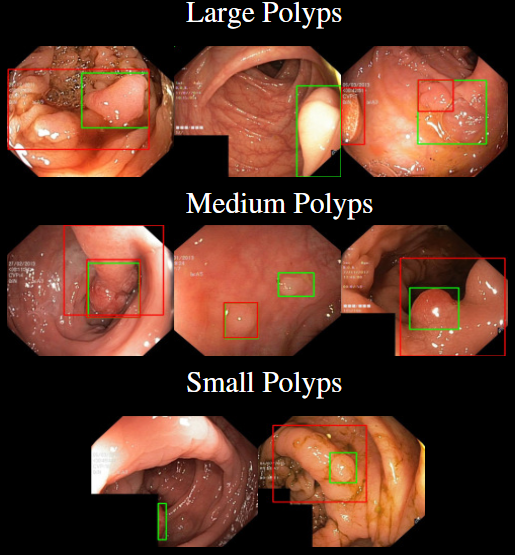}
\caption{Prediction error on large, medium and small. Green and red boxes represent the true and predicted bounding boxes respectively.}\label{fig9}
\end{figure}

On analyzing the failure cases, we found that for some polyps, the YOLOv8 model produces overly extended boxes, capturing adjacent tissue occasionally. In Fig. \ref{fig9}, the first row shows that for both large and medium polyps, YOLO-LAN sometimes predicts either a larger area or misses some polyps. For small polyps, one of the polyp missed is masked due to confidential information overlap and another one is detected along with adjacent tissue. Despite its minor errors shown above, YOLO-LAN outperforms all the existing results on polyp detection.

\section{Discussion and Conclusion}

Polyp regions can vary greatly in size, shape, and texture, while it is usually easier to manually identify larger bulkier polps, smaller lesions are often most easily missed. We therefore propose the YOLO-LAN pipeline, where three specific final design decisions were made: i) negative samples provide a closer approximation of real-world clinical scenarios, where many frames will have no polyps; ii) a loss function based on M2IoU that increases localization precision of predicted bounding box; and iii) planned augmentations to introduce diversity in datasets. 

In contrast to much of the prior work that takes as input only polyp-containing images, we took a different approach in explicitly including non-polyp from PolypGen2021 and Kvasir into the training process. As the final product, the reduced negatives at controlled proportions (10\% and 20\%) provided considerable reduction in false positives that were learnt from accepting non-polyp regions as clearly non-polyp. The prepared augmentations (spatial transformations, blur distortions, and composite transformations) resulted in data expansion from 700 to almost 7000 training images, with proportionally well balanced negatives. This augmentation process presented not only robust, non-specific effects to visual variance learned in non-polyp images, but simulating clinical imaging conditions.

Our results on benchmark dataset like Kvasir-Seg achieve state-of-the-art performance, with significant improvements in mAP$_{50:95}$, and improved localization of polyps in size. For example, YOLOv8 variants were able to achieve precision of 0.9597, recall of 0.9395, and F1-score of 0.9400, mAP$_{50}$ of 0.9540, and mAP$_{50:95}$ of 0.8487 while YOLOv12 improved with an precision of 0.9688, recall of 0.9519, F1-score of 0.9464, mAP$_{50}$ of 0.9619, and mAP$_{50:95}$ of 0.8587 on the Kvasir-seg dataset. Similar performance gains were achieved on the BKAI-IGH NeoPolyp dataset demonstrating the generalizability of our approach. The gains in mAP$_{50:95}$ are especially significant, as it is a stricter and more clinically relevant metric, it reinforces the accurate potential of our detection pipeline. Furthermore, the strong performance of YOLOv8/12 small variants also supports their feasibility for real-time operations on resource-constrained systems. The method's accuracy can therefore translate into clinically viable changes.

Despite YOLO-LAN yielding compelling results, there are still open problems. Firstly, future work should engage temporal modeling to allow for frame-to-frame attribution of continuity and allow the model to learn from previous frames to reduce false detections from video streams. Second, while we utilized negative samples to create a more realistic training set with respect to polyp-abundant versus polyp-rare conditions, we still do not know the best ratio of non-polyp frames that are needed during polyp training. Finally, it will be paramount to evaluate the full pipeline on large, multi-center datasets and then prospective clinical trials to confirm generalizability and true clinical use.

\end{document}